\documentclass[10pt,twocolumn,letterpaper]{article}

\usepackage{cvpr}              

\usepackage[table]{xcolor}
\usepackage{url}
\usepackage{xspace}
\usepackage{graphicx}
\usepackage{subcaption}
\usepackage{listings}
\usepackage{xcolor}
\usepackage{titletoc}
\usepackage{amsmath}
\usepackage[font={small}]{caption}
\usepackage[export]{adjustbox}
\usepackage{bm}
\usepackage{booktabs}
\usepackage{balance}
\usepackage{color}
\usepackage{dsfont}
\usepackage{esvect}
\usepackage{float}
\usepackage{graphicx}
\usepackage{pifont}
\usepackage{import}
\usepackage{mathtools}
\usepackage{multirow}
\usepackage{stackengine}
\usepackage{stmaryrd}
\usepackage{systeme}
\usepackage{soul}
\usepackage{tabularx}
\usepackage{url}
\usepackage{tikz}
\usetikzlibrary{tikzmark}
\usetikzlibrary{calc}
\usepackage[normalem]{ulem}
\usepackage{enumitem}
\usepackage{algorithm2e}
\usepackage{scalerel}
\usepackage{wrapfig}
\usepackage{bxcoloremoji}
\usepackage{lipsum}
\usepackage{tocloft}
\usepackage{cuted}
\usepackage[dvipsnames]{xcolor}
\usepackage[accsupp]{axessibility} 

\definecolor{citecolor}{HTML}{0071bc}
\usepackage[pagebackref,breaklinks,urlcolor=blue,colorlinks,citecolor=citecolor,bookmarks=false]{hyperref}

\definecolor{codeblue}{rgb}{0.25,0.5,0.5}
\definecolor{nvgreen}{rgb}{0.92, 0.97, 0.85}

\definecolor{navyblue}{HTML}{0071BC}
\definecolor{hotpink}{HTML}{FF0080}

\definecolor{myyellow}{HTML}{FFD966}
\definecolor{myred}{rgb}{1, 0.9, 0.9}
\definecolor{mygray}{gray}{0.95}

\newcommand{\myparagraph}[1]{\vspace{-0.2mm}\noindent\textbf{#1}}

\newlength\paramargin
\newlength\figmargin
\newlength\subfigmargin
\newlength\secmargin
\newlength\subsecmargin
\newlength\tabmargin
\newlength\eqmargin

\SetKwComment{Comment}{\color{green!50!black}\# }{}

\SetKwProg{Function}{def}{:}{}

\SetKwProg{For}{for}{:}{}
\SetKwProg{If}{if}{:}{}

\makeatletter
\newcount\my@rainbow@cnt
\def\rainbowtext#1{%
  \my@rainbow@cnt=0\relax
  \rainbow@text#1\@nil
}
\def\rainbow@text#1#2\@nil{%
  \ifx#1\relax\else
    \ifcase\my@rainbow@cnt
      \color{Red}%
    \or
      \color{Orange}%
    \or
      \color{Yellow}%
    \or
      \color{Green}%
    \or
      \color{Blue}%
    \or
      \color{Violet}%
    \fi
    #1%
    \global\advance\my@rainbow@cnt by 1\relax
    \ifnum\my@rainbow@cnt>5\global\my@rainbow@cnt=0\fi
    \rainbow@text#2\@nil
  \fi
}
\makeatother

\def\method{GR3D\xspace}
\def\model{GR3D-8B\xspace}

\definecolor{cvprblue}{rgb}{0.21,0.49,0.74}

\def\paperID{} 
\def\confName{CVPR}
\def\confYear{2026}

\title{Grounded 3D-Aware Spatial Vision-Language Modeling}

\author{
        An-Chieh Cheng$^{1,3}$\footnotemark[1] \hspace{0.2em}
        Yang Fu$^{1}$ \hspace{0.1em}
        Yatai Ji$^{3}$ \hspace{0.1em}
        Ligeng Zhu$^{3}$ \hspace{0.1em}
        Guanqi Zhan$^{3}$ \hspace{0.1em}
        Zhuoyang Zhang$^{2,3}$ \\[0.4em] 
        Zhaojing Yang$^{1}$ \hspace{0.1em}
        Song Han$^{2,3}$ \hspace{0.1em} 
        Yao Lu$^{3}$ \hspace{0.1em}
        Pavlo Molchanov$^{3}$ \hspace{0.1em}
        Vidya Nariyambut Murali$^{3}$ \hspace{0.1em}
        Jan Kautz$^{3}$ \\[0.4em]
        Xiaolong Wang$^{1}$ \hspace{0.1em}
        Hongxu Yin$^{3}$ \hspace{0.1em}
        Sifei Liu$^{3}$ \\[0.4em]
        $^1$UCSD \quad
        $^2$MIT \quad
        $^3$NVIDIA
}

\begin{document}



\twocolumn[{
      \vspace{-2em}
      \maketitle
      
      \vspace{-3em}
      \begin{center}
      \url{https://www.anjiecheng.me/gr3d}
      \end{center}
      \begin{center}
        \centering
        \includegraphics[width=\textwidth]{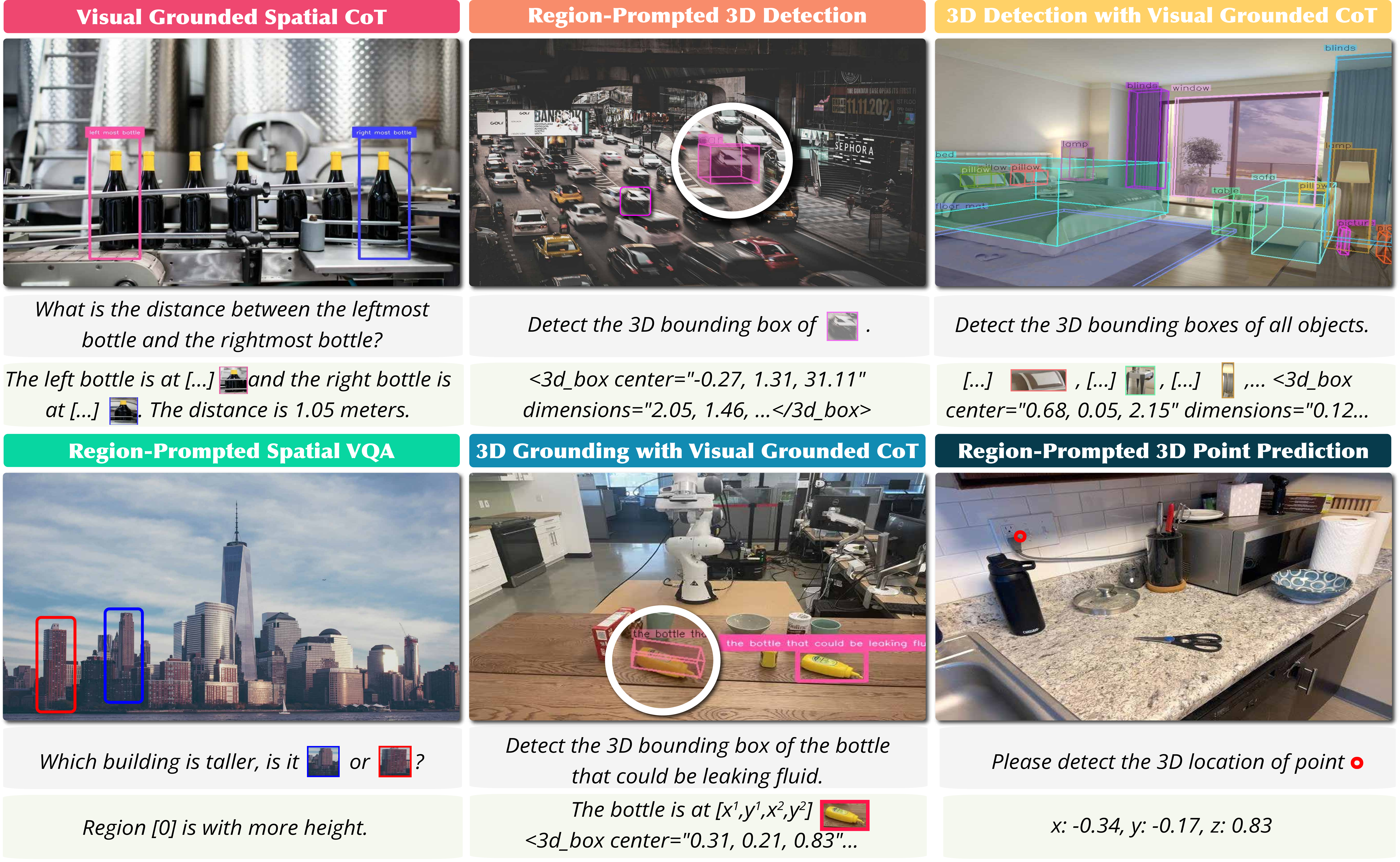}
        \vspace{-1.5em}
        \captionof{figure}{
          \textbf{Overview.} GR3D bridges 2D pixel-space and 3D metric-space by integrating multiple grounding capabilities into visual chain-of-thoughts, enabling complex spatial understanding through grounded 2D perception followed by 3D inference.
        }\vspace{-1.5mm}
        \label{fig:teaser}
    \end{center}
}]

\footnotetext[1]{Work done during an internship at NVIDIA.}
\vspace{-0.5em}
\begin{abstract}

We present GR3D, a spatial vision language model equipped with three complementary grounding capabilities—explicit 2D grounding, implicit 2D grounding, and monocular 3D grounding—within a single framework. GR3D introduces an implicit grounding mechanism that identifies entity mentions during generation and inserts the corresponding region tokens into the text stream, allowing the model to reference visual evidence on the fly when producing spatial chain-of-thought responses. In parallel, a region-prompted monocular 3D grounding design predicts 3D bounding boxes in the camera view from grounded region queries, supported by intrinsic-aware normalization and dense geometric supervision. Together, these grounding capabilities enable GR3D to decompose complex spatial understanding problems into grounded 2D perception followed by 3D inference. GR3D achieves consistent improvements across grounded and non-grounded spatial benchmarks, demonstrating grounding as an effective inductive bias for strengthening spatial understanding in VLMs.
These grounding capabilities collectively enhance general spatial understanding beyond the grounding task itself. 
\end{abstract}\vspace{-5mm}    
\section{Introduction}
\label{sec:intro}

Vision–language models (VLMs) have rapidly evolved into general-purpose perception–language systems
\cite{liu2024visual,lin2024vila,nvila,peng2023kosmos,team2023qwen,internvl,openai2023chatgpt,team2023gemini},
capable of understanding scenes, following open-ended instructions, and supporting diverse multimodal tasks.
As these models begin to serve as the core of embodied agents that must act, manipulate, and navigate in the physical world
\cite{szot2021habitat,grauman2024ego,brohan2023rt2,robocasa2024,cheng2024navila,yang2025egovla,gr00tn1_2025,intelligence2025pi05,molmoact2025},
their spatial competence becomes crucial. Embodied intelligence requires models not only to recognize what is present, but also to understand where objects are and how they are arranged in space—capabilities essential for grounding language into actions such as where to reach, step, or orient
\cite{shridhar2022cliport,brohan2023rt1,sermanet2024robovqa}.
Without reliable spatial grounding, the link between high-level instructions and physical interaction remains brittle, limiting the scalability of VLMs toward real-world embodied perception and control.

Rapid progress in spatial VLMs has substantially advanced 2D spatial understanding and even 3D perception
\cite{srgpt,cheng20253d,zhou2025roborefer,RoboBrain2.0,yuan2024robopoint,vst,llava3d,video3dllm,huang20253d}.
Yet grounding—the ability to reliably associate linguistic mentions with concrete visual regions and connect 2D evidence with 3D structure—remains limited.
Two challenges, in particular, are under-addressed.
(i) Implicit 2D grounding is scarce: most systems support explicit “point to X” grounding but lack mechanisms or data for automatically detecting entities mentioned in free-form text and integrating their corresponding visual evidence during generation. Constructing such supervision is difficult, as it requires aligning textual mentions to latent visual regions and interleaving region information into the language stream.
(ii) Monocular 3D grounding is inherently ill-posed: from a single view, object scale, depth, and intrinsics are entangled, and 3D prediction requires first identifying which instance the text refers to before estimating its 3D extent and pose. Existing approaches often bypass this intermediate localization step~\cite{geminirobotics}, rely on multi-view supervision~\citep{dai2017scannet}, or are limited by the scarcity of 3D box annotations~\cite{brazil2023omni3d}.

To address these limitations, we introduce \textbf{(GR3D)}, a spatial VLM that integrates grounding as a core mechanism for learning spatial representations. GR3D jointly supports three complementary grounding capabilities within a unified architecture: \textit{explicit 2D grounding}, which predicts object regions through the language head in a structured textual format; \textit{implicit 2D grounding}, which links linguistic mentions to visual evidence through dynamic region insertion; and \textit{monocular 3D grounding}, which extends region understanding into 3D by predicting bounding boxes and camera intrinsics under dense geometric supervision. Together, these mechanisms establish a fine-grained alignment between language, image regions, and geometry, enabling consistent 2D and 3D spatial reasoning.

While explicit 2D grounding predicts the location of queried objects, it cannot handle free-form reasoning where spatial cues are implicit. Real-world spatial queries—\eg, describing relations, distances, or navigation targets—require first recognizing and localizing the entities mentioned before reasoning about the query itself. GR3D bridges this gap with an implicit 2D grounding mechanism that performs \textit{streaming region insertion}: as the model generates responses, it dynamically predicts the visual region corresponding to each mentioned entity, encodes the region into a token, and injects it directly into the ongoing language stream. This enables reasoning to evolve directly over grounded visual evidence, yielding coherent spatial predictions without any separate detection phase.


Inferring 3D structure from a single view introduces both linguistic and geometric ambiguities, such as determining which instance a description refers to and estimating its depth, scale, and pose without multi-view cues. GR3D addresses these challenges through a region-prompted 3D grounding formulation: each grounded 2D region is treated as a query for 3D inference, supported by intrinsic-aware normalization and dense geometric supervision derived from depth estimation. This design aligns semantic localization and geometric prediction within a consistent camera-view framework, enabling the model to infer coherent 3D structure directly from grounded 2D evidence and to generalize across diverse scenes and viewpoints.
Crucially, by receiving region tokens produced by implicit 2D grounding, the 3D predictor naturally plugs into CoT-driven reasoning—allowing the model to first resolve “which object” via grounded language generation and then infer “what 3D structure” for that object. This decomposition makes monocular 3D grounding applicable to both instance-level referring tasks and open-set category-level 3D detection.

Integrating explicit 2D grounding, implicit 2D grounding, and monocular 3D grounding positions GR3D as a flexible spatial understanding framework spanning 2D/3D and single-/multi-view settings. Through this grounding-centered formulation, the model learns to localize, reference, and reason over spatial structure in a unified manner. Implicit grounding enhances CoT accuracy and spatial consistency on CVBench~\cite{tong2024cambrian}, ERQA~\cite{geminirobotics} and SAT~\cite{ray2024sat}, while region-prompted 3D grounding with dense point supervision achieves state-of-the-art performance on Omni3D. Moreover, we observe key insights:
(i) grounding improves general spatial understanding even without explicit localization;
(ii) dense geometric supervision provides scalable structure cues;
(iii) combining implicit grounding with region-prompted 3D inference unlocks a versatile decomposition pipeline that supports referring-instance 3D grounding, category-level 3D detection, and multi-object scene grounding.
Together, these results show that embedding grounding within the model architecture strengthens both spatial perception and grounded reasoning.

\begin{figure*}[!t]
\center
\includegraphics[width=1\textwidth]{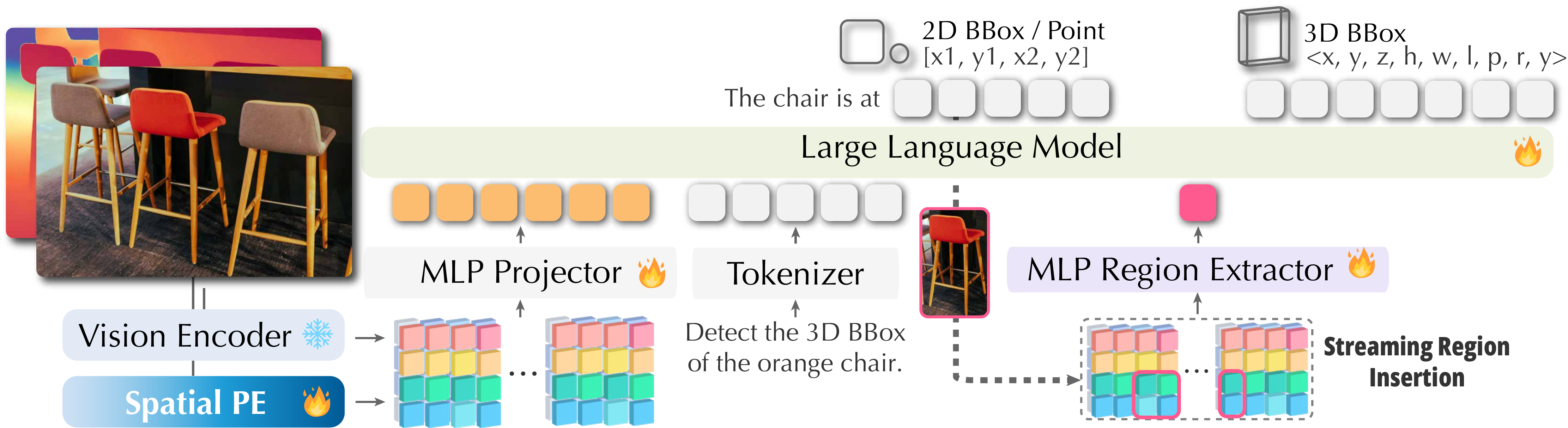}
\vspace{-1.5em}
\caption{\label{fig:method}Method overview. GR3D builds on Region-VLMs by adding streaming region insertion for visual Chain-of-Thought reasoning. During CoT, the model repeatedly predicts a region, extracts its visual embedding, and reinserts a region token into the text sequence, enabling step-by-step spatial reasoning with dynamically refreshed visual cues.}
\vspace{-1em}
\end{figure*}

\section{Method}
\label{sec:method}
GR3D is designed to address two major limitations of current Spatial VLMs: the lack of an implicit grounding mechanism that allows models to automatically associate linguistic mentions with visual evidence during reasoning, and the difficulty of performing monocular 3D grounding from a single image with entangled depth and scale cues. To overcome these, we first construct a \textbf{foundational Spatial VLM} (Sec.~\ref{subsec:camera_relative}) that provides geometry-aware features for both single- and multi-view inputs. Building on this foundation, we introduce \textbf{explicit} and \textbf{implicit 2D grounding} (Sec.~\ref{subsec:grounding_in_2d}) to link linguistic expressions with visual evidence, and extend them to \textbf{monocular 3D grounding} through region prompts, intrinsic normalization, and dense geometric supervision (Sec.~\ref{sec:ground}). Finally, we describe our \textbf{data construction pipeline} (Sec.~\ref{sec:data}) that generates large-scale implicit grounding annotations and balanced 2D–3D supervision to facilitate training.

\subsection{Foundational Spatial VLM}
\label{subsec:camera_relative}
\noindent\textbf{Objective.}
We follow the design principle of SR-3D~\cite{cheng20253d} to construct a foundational Spatial VLM based on the NVILA-8B-Lite~\cite{nvila} architecture.  
This model provides a unified spatial representation that supports both single- and multi-view spatial understanding, serving as the base for subsequent grounding modules.  
At this stage, no grounding capability is included; the focus is on building a geometry-aware representation layer compatible with language reasoning.

\noindent\textbf{Single-view Setup.}
The base NVILA encoder extracts dense visual tokens from an RGB image for single-view inputs.
To make these tokens spatially aware, we augment them with 2D positional embeddings derived from their pixel coordinates and relative depth cues.
Each visual token therefore carries both appearance and geometric context.
Unlike language tokens, which are processed sequentially, these enriched visual tokens retain their spatial arrangement within the image grid.
They are passed through the multimodal projector before being fed to the language model.
This projection preserves spatial locality while remaining compatible with NVILA’s multimodal fusion pipeline.

In addition, we preserve the region-prompt design used in SR-3D: specific image regions can be encoded as individual query tokens by pooling features within a given bounding box.
This structure allows downstream modules to reference localized spatial content directly, maintaining full alignment with the NVILA token structure and positional hierarchy.
Overall, the single-view formulation provides a strong spatially-structured feature space for both region-level interaction and text-aligned representation.

\noindent\textbf{Multi-view Setup.}
Our framework naturally extends from single-view to multi-view inputs by embedding all image tokens with depth- and pixel-based positional cues in a unified spatial feature space. The first view is processed exactly as in the single-view case, and all subsequent views are transformed into the first-frame coordinate system. Please see the supplementary materials for our multi-view results.


\subsection{Grounding in the 2D Plane}
\label{subsec:grounding_in_2d}
Grounding on the 2D plane aims to teach the model to associate linguistic mentions with localized image evidence. We introduce both explicit and implicit forms of grounding, designed to strengthen the spatial reasoning capacity of the vision--language model.

\subsubsection{Explicit 2D Grounding}
For explicit 2D grounding, we adopt a simple and general formulation. Given a natural-language instruction, the model predicts 2D bounding boxes directly in HTML-style textual format (\eg, \texttt{<bbox>[x$_1$, y$_1$, x$_2$, y$_2$]</bbox>}), using its standard language generation head without any additional detection branch. This unified design integrates grounding seamlessly into the vision--language interface, without introducing task-specific architectural components.

\subsubsection{Implicit 2D Grounding}
Consider a global spatial reasoning query such as:
\textit{``In the kitchen, how far is the second bottle on the shelf from the small brown teddy bear on top of the washing machine in the laundry room?''}
Traditional spatial VLMs attempt to answer such questions directly from global image features, relying on large-scale question--answer pairs to memorize spatial relationships. However, this departs from how humans perceive scenes: we first identify where each mentioned object is before reasoning about their relations. Our implicit grounding mechanism explicitly introduces this intermediate step of \textit{entity localization during generation}, aligning the model's behavior with human visual reasoning.

\noindent\textbf{Streaming Region Insertion.}
Given an input instruction, the model generates its response in a chain-of-thought (CoT) fashion. When an entity (\eg, ``the second bottle on the shelf'') is mentioned, the model first predicts its corresponding 2D bounding box coordinates, \eg, $[x_1, y_1, x_2, y_2]$. Immediately after, the corresponding image region is encoded through the region encoder, and its embedding---a region token---is inserted directly into the text stream at that position. The generation then continues conditioned on both the textual and visual context. The same procedure repeats for subsequent entities, producing a temporally aligned reasoning trajectory that alternates between language and vision.

\noindent\textbf{Training and Inference Paradigm.} During training, the bounding box coordinates are directly predicted by the language head and optimized through teacher forcing, as they are treated as part of the textual output sequence. Once the coordinates are produced, the corresponding region token---derived from the ground-truth region---is inserted into the generation stream. This token is detached from the computation graph (i.e., no gradient flows through it) but serves as a strong conditional cue for subsequent token prediction. 
During inference, the process becomes fully autoregressive. The model first predicts coordinates, then encodes the predicted region to obtain its embedding, which is inserted back into the ongoing sequence before the next generation step. The subsequent reasoning, such as relational comparison or distance estimation, is thus conditioned on both the textual context and dynamically inserted region evidence.

\begin{table*}[tbp]
\footnotesize
    \centering

    \vspace{-3mm}
    \scalebox{0.97}{
    \setlength{\tabcolsep}{1pt}
    \begin{tabular}{lcccccccccccccc}
    \toprule
    \multirow{3}{*}{Method}
    & \multicolumn{2}{c}{\small\textsc{Sun-rgbd}~\cite{sunrgbd}}
    & \multicolumn{2}{c}{\small\textsc{Arkitscenes}~\cite{arkitscenes}}
    & \multicolumn{2}{c}{\small\textsc{Objectron}~\cite{objectron}}
    & \multicolumn{2}{c}{\small\textsc{Hypersim}~\cite{hypersim}}
    & \multicolumn{2}{c}{\small\textsc{Kitti}~\cite{kitti}}
    & \multicolumn{2}{c}{\small\textsc{Nuscenes}~\cite{nuscenes}}
    & \multirow{3}{*}{${\rm AP_{3D}} \uparrow$} \\
    \cmidrule(lr){2-3} \cmidrule(lr){4-5} \cmidrule(lr){6-7} \cmidrule(lr){8-9}
    \cmidrule(lr){10-11} \cmidrule(lr){12-13}
    & ${\rm AP_{15}} \uparrow$ & ${\rm mAP} \uparrow$
    & ${\rm AP_{15}} \uparrow$ & ${\rm mAP} \uparrow$
    & ${\rm AP_{15}} \uparrow$ & ${\rm mAP} \uparrow$
    & ${\rm AP_{15}} \uparrow$ & ${\rm mAP} \uparrow$
    & ${\rm AP_{15}} \uparrow$ & ${\rm mAP} \uparrow$
    & ${\rm AP_{15}} \uparrow$ & ${\rm mAP} \uparrow$
    & \\
    \midrule
    \multicolumn{14}{c}{\textit{Vision Specialist Models}} \\
    \midrule

    ImVoxelNet~\cite{rukhovich2022imvoxelnet}
    & - & - & - & - & - & - & - & - & - & - & - & - & 9.4 \\

    SMOKE~\cite{liu2020smoke}
    & - & - & - & - & - & - & - & - & - & - & - & - & 10.4 \\

    Cube R-CNN~\cite{brazil2023omni3d}
    & - & 15.33
    & - & 41.73
    & - & 50.84
    & - & 7.48
    & - & 32.50
    & - & 30.06
    & 23.26 \\

    OVMono3D~\cite{ovmono3d}\textsubscript{\emph{w/ Cube R-CNN}}
    & - & 15.20
    & - & 41.60
    & - & 58.87
    & - & 7.75
    & - & 25.45
    & - & 24.33
    & 22.98 \\

    DetAny3D~\cite{detany3d}\textsubscript{\emph{w/ Cube R-CNN}}
    & 26.62 & 18.96
    & 59.55 & 46.13
    & 72.51 & 54.42
    & 11.43 & 7.17
    & 44.28 & 31.61
    & 41.01 & 30.97
    & 24.92 \\

    \midrule
    \multicolumn{14}{c}{\textit{Vision Language Models}} \\
    \midrule


    Qwen3-VL-4B~\cite{qwen3vl}
    & 28.28 & 17.60
    & 63.97 & 46.33
    & 61.60 & 43.13
    & 11.56 & 6.44
    & 17.39 & 11.25
    & 7.48 & 4.89
    & - \\

    Qwen3-VL-8B~\cite{qwen3vl}
    & 28.28 & 17.77
    & 62.32 & 45.23
    & 61.63 & 43.59
    & 11.62 & 7.23
    & 5.23 & 3.32
    & 11.52 & 7.56
    & - \\



    \rowcolor{myyellow!70}
    \textbf{\model} (Ours)
    & \bf43.49 & \bf31.64
    & \bf67.49 & \bf52.52
    & \bf71.68 & \bf54.32
    & \bf16.42 & \bf10.87
    & \bf22.18 & \bf14.75
    & \bf22.98 & \bf16.59
    & \bf25.40 \\
    
    \bottomrule
    \end{tabular}
    }
    \vspace{-1em}
\caption{
Comparison on the Omni3D~\cite{brazil2023omni3d} benchmark between \method, vision specialists, and recent VLMs. We report ${\rm AP_{15}}$ and ${\rm mAP}$ for each dataset domain. \method outperforms all recent VLMs and vision specialists, especially on the indoor domain.
}
    \label{tab:det3d}
    \vspace{-2em}
\end{table*}

\begin{table}[tbp]
\footnotesize

    \centering
    \scalebox{0.97}{
    \setlength{\tabcolsep}{2.75pt} 
    \begin{tabular}{lcccccc} 
    \toprule
    Method
    & ${\rm AP^{sun}_{2D}}$
    & ${\rm AP^{ark}_{2D}}$
    & ${\rm AP^{obj}_{2D}}$
    & ${\rm AP^{hyp}_{2D}}$
    & ${\rm AP^{kit}_{2D}}$
    & ${\rm AP^{nus}_{2D}}$ \\
    
    \midrule
    Cube R-CNN~\cite{brazil2023omni3d}
    & 15.07 & 40.22 & 49.24 & 11.05 & 36.14 & 34.64  \\
    
    
    Qwen3-VL-8B~\cite{qwen3vl}
    & 8.06 & 22.44 & 30.06 & 3.08 & 1.54 & 2.56 \\
    
    \rowcolor{myyellow!70}
    \textbf{\model} (Ours)
    & \textbf{38.86} & \textbf{46.17} & \textbf{51.66} & \textbf{28.53} & \textbf{20.49} & \textbf{22.16} \\
    \bottomrule
    \end{tabular}
    } 
    \vspace{-1em}
    \caption{
    2D detection results on the Omni3D benchmark. We report the mean Average Precision (mAP) for each dataset domain.
    }
    \label{tab:indomain_ap2d}
    \vspace{-2em}
\end{table}

\noindent\textbf{Comparison and Interpretation.}
Our stream-based grounding can be viewed abstractly as analogous to a two-step process, \ie, first grounding entities with a VLM, and then performing region-conditioned reasoning with a spatial VLM with a region encoder equipped. Unlike this staged formulation, our approach unifies both phases in a single generative stream. The model learns \emph{when} and \emph{what} to ground based on linguistic context, and its reasoning naturally unfolds on grounded evidence without explicit stage transitions. This results in a fluid, interpretable reasoning process that tightly couples perception and cognition while avoiding the discontinuities of discrete grounding modules.

\subsection{Monocular 3D Grounding via Region Prompt}\label{sec:ground}
Monocular 3D grounding aims to enable single-view models to infer 3D structure from natural language and visual cues. This task faces two major challenges. First, linguistic ambiguity: textual references often under-specify which instance is being mentioned, requiring the model to implicitly identify the target entity before 3D reasoning. Second, geometric ambiguity: the coupling between object scale, depth, and camera intrinsics makes single-view estimation inherently uncertain. We address these through several components below that align semantic localization and geometric inference within a unified generative framework.

\noindent\textbf{Region-prompt Formulation.}
Given a localized 2D region, the model treats this region as a spatial query for 3D reasoning. The region’s visual features are pooled and encoded into a region token, which is fused into the text stream to guide 3D box prediction. Since the model already possesses implicit 2D grounding capability, this step focuses solely on extending that capacity from 2D to 3D—mapping a grounded region to its corresponding 3D representation. This formulation simplifies 3D grounding by conditioning inference on a given region, enabling the model to estimate position, scale, and orientation directly without performing explicit multi-step localization.

\noindent\textbf{3D Box Representation.}
Each 3D bounding box is expressed in a unified, language-based format compatible with 2D HTML-style outputs, eliminating the need for task-specific heads. The box is parameterized by its center $(x_c, y_c, z_c)$, size $(w, h, l)$, and orientation $(\theta_p, \theta_r, \theta_y)$, where $(\theta_p, \theta_r, \theta_y)$ are \emph{normalized} Euler angles (pitch/roll/yaw). To ensure consistency across datasets, we standardize orientations by selecting the rotation variant that minimizes the angular deviation between the local PCA axes of the region and the global coordinate axes $(X,Y,Z)$—that is, the variant closest to the identity basis rather than a mirrored alternative. This compact decomposition makes the representation transferable: the center term aligns naturally with depth-based supervision (see below), while the dimension and rotation terms capture view-invariant geometry. The format promotes stability, interpretability, and seamless integration into the generative language interface.

\noindent\textbf{Intrinsic Normalization.}
To mitigate scale and depth ambiguity, we introduce an intrinsic-aware normalization strategy that rescales images according to focal length, yielding a consistent field of view across datasets. Concretely, given focal length $f_x$, we normalize the spatial scale by $W' = \tfrac{1000}{f_x} \cdot W$ and $H' = \tfrac{1000}{f_x} \cdot H$, aligning the apparent object size in the feature space and supporting robust 3D inference without explicitly regressing intrinsics.

\begin{table*}[!th]
\footnotesize
    \vspace{-3mm}
    \centering
    \scalebox{0.97}{
    \setlength{\tabcolsep}{1.2pt}
    \begin{tabular}{l ccc cccc c c c c c c c c} 
    \toprule
    \multirow{3}{*}{Method}
    & \multicolumn{11}{c}{\small\textsc{Spatial}}

    & \multicolumn{4}{c}{\small\textsc{General}} \\
    \cmidrule(lr){2-12} \cmidrule(lr){13-16}
    & \multicolumn{3}{c}{BLINK~\cite{fu2024blink}}
    & \multicolumn{4}{c}{CVBench~\cite{tong2024cambrian}}
    & \multirow{2}{*}{RWQA~\cite{rwqa}}
    & \multirow{2}{*}{ERQA~\cite{geminirobotics}}
    & \multirow{2}{*}{SAT~\cite{ray2024sat}}
    & \multirow{2}{*}{EMB~\cite{embspatial}}
    & \multirow{2}{*}{ChartQA~\cite{masry2022chartqa}}
    & \multirow{2}{*}{MME~\cite{mme}}
    & \multirow{2}{*}{POPE~\cite{pope}}
    & \multirow{2}{*}{AI2D~\cite{kembhavi2016diagram}} \\
    \cmidrule(lr){2-4} \cmidrule(lr){5-8}
    & Dep. & Spa. & Avg.
    & Rel. & Dep. & Dis. & Avg.
    & & & & & & & & \\
    \midrule
    
    NVILA-Lite-8B
    & 73.38 & 79.72 & 50.51 & 93.38 & 92.83 & 91.00 & 86.31 & 65.35 & 36.25 & 62.60 & 68.90 & 84.80 & 1692 & 88.10 & 91.01 \\
    
    
    \midrule
    \rowcolor{myyellow!70}
    \textbf{\model} (Stage 1)
    & \textbf{87.90} & \textbf{83.21} & \textbf{54.35} & \textbf{96.92} & \textbf{98.16} & \underline{95.50} & \underline{87.23} & \textbf{68.75} & \textbf{40.25} & \textbf{76.00} & \textbf{81.01} & \textbf{84.48} & \textbf{1656} & \textbf{88.23} & \textbf{91.81} \\
    \textbf{\model}  (Stage 2)
    & \textbf{87.90} & \underline{80.41} & \underline{53.26} & \underline{96.46} & \underline{98.00} & \textbf{96.00} & \textbf{87.26} & \underline{65.23} & \underline{38.50} & \underline{70.60} & \underline{77.58} & \underline{84.00} & \underline{1626} & \underline{87.00} & \underline{91.54} \\
    \bottomrule
    \end{tabular}
    } 
    \vspace{-1em}
    \caption{
    Performance comparison on general visual question answering and spatial reasoning benchmarks.
    }
    \label{tab:spatial_benchmarks_condensed}
    \vspace{-2em}
\end{table*}

\noindent\textbf{Points and Direct Grounding Supervision.}
We supervise monocular 3D grounding with complementary signals beyond sparse 3D-box labels. \emph{(i) Region$\rightarrow$3D:} when a 2D box is available, the model predicts its 3D box directly from the region prompt. \emph{(ii) Pure text$\rightarrow$3D:} when no 2D box exists, the model localizes the mentioned entity via its built-in textual grounding and regresses its 3D box, enabling coverage of text-only data. In addition, we construct an auxiliary dense region-to-3D supervision: from ground-truth or predicted depth maps, we randomly sample valid surface points per image (\eg, 100 per image) and train the model to predict their 3D coordinates conditioned on the corresponding region prompt. This depth-driven signal scales supervision well beyond limited 3D-box annotations. Finally, to tolerate modest grounding noise, we apply lightweight 2D bounding-box augmentation (jitters in size and location), improving robustness while preserving semantic locality.

Together, region-prompt grounding, structured 3D box representation, intrinsic normalization, and scalable training signals address both linguistic and geometric ambiguities of monocular 3D grounding. These components jointly provide a camera-relative spatial understanding that generalizes across datasets and supports future extensions to multi-view and embodied reasoning tasks.

\subsection{Data Construction and Composition}\label{sec:data}
\noindent\textbf{Data Construction for Grounding.} To construct the implicit grounding corpus, we start from RefSpatial~\cite{zhou2025roborefer}, which includes 2D samples from OpenImages~\cite{openimages}, 3D video data from CA-1M~\cite{ca1m}, and synthetic scenes. RefSpatial contains diverse image–text pairs, but it lacks region-level annotations for all the mentioned entities. To obtain them, we use Florence-2~\cite{xiao2024florence} to generate candidate 2D bounding boxes and class labels for each textual mention, producing dense but noisy region annotations.

We then refine these annotations through a VLM for verification and a rephrasing pipeline. This process (i) verifies one-to-one alignment between textual mentions and detected regions, removing unmatched or ambiguous cases, and (ii) rewrites generic class names into concise, instance-level descriptions based on image context. The resulting corpus provides high-quality implicit grounding supervision that links textual mentions to corresponding visual evidence with precise instance semantics.

For explicit grounding, we augment samples that contain ground-truth boxes by generating short instance-level referring expressions with a vision–language model and validating their existence in the image. Only verified matches are retained. Together, these procedures yield reliable implicit and explicit grounding data, while depth, point, and 3D-box supervision follow the setup in Sec.~\ref{sec:ground}.

\noindent\textbf{Data Composition and Distribution.}
Our training data is composed of publicly available sources: 97K grounded CoT samples, 780K 3D detection samples from Omni3D~\cite{brazil2023omni3d} and EmbodiedScan~\cite{wang2024embodiedscan}, and 272K pointmap reconstruction samples from DepthLM~\cite{cai2025depthlm}. We do not use any proprietary or in-house data and the scale of our 3D detection data is comparable to prior works such as VST~\cite{vst}, ensuring performance gains are not simply due to data size.

\begin{figure}[!t]
\centering
\footnotesize
\begin{minipage}[c]{0.2\textwidth} 
\centering
\footnotesize
\scalebox{0.95}{
\setlength{\tabcolsep}{1.2pt} 
\begin{tabular}{lc}
    \toprule
    Methods & Acc. (\%) \\ \midrule
    \color[HTML]{969696}Human & \color[HTML]{969696}98.3  \\ \midrule
    GPT-4V-Turbo~\cite{openai2023chatgpt} & 66.9  \\ 
    GPT-4o~\cite{openai2024gpt4o} & 64.5  \\ 
    LLaVA-v1.5-7B-xtuner~\cite{2023xtuner} & 50.8 \\
    CogVLM-7B~\cite{wang2023cogvlm} & 50.8 \\
    LLaVA-v1.5-7B~\cite{liu2024improved} & 51.6 \\
    LLaVA-InternLM2-7B~\cite{cai2024internlm2} & 52.4  \\ 
    SpatialRGPT-8B*~\cite{srgpt} & 87.9  \\ 
    SR3D-8B*~\cite{cheng20253d} & 90.3  \\
    \rowcolor{myyellow!70}
    \textbf{\model} (Ours) & \bf94.4 \\
    \bottomrule
\end{tabular}
}
\end{minipage}
\hfill 
\begin{minipage}[c]{0.23\textwidth}
\centering
\includegraphics[width=3.75cm,height=4.5cm]{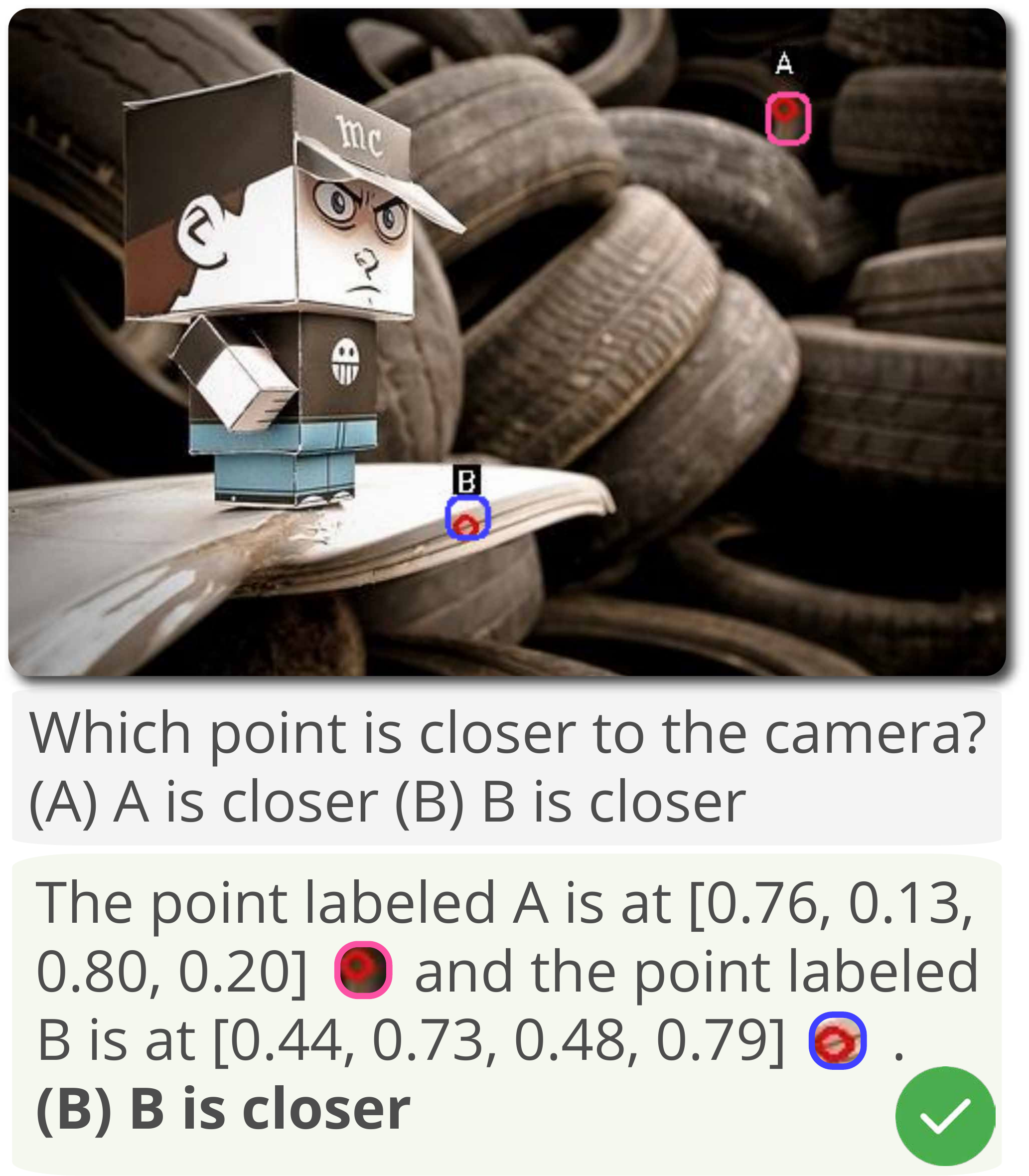} 
\end{minipage}
\vspace{-1em}
\caption{
Results on the BLINK-Depth benchmark for point-level region spatial understanding.  
Left: comparison with VLM baselines.  
Right: visualization of one sample.  
Our method surpasses prior Region-VLMs (*), which require manual annotated masks.
}
\label{fig:table_figure_shared}
\vspace{-2em}
\end{figure}

\section{Experiments}
\label{sec:exp}
In this section, we begin by describing the implementation details (Sec.~\ref{sec:exp:detail}), including the training stages and datasets used. We then present the main results of our model, highlighting its 3D detection performance in Sec.~\ref{sec:exp:det3d}. Next, we assess whether the model preserves its general VLM and spatial capabilities in Sec.~\ref{sec:exp:vqa}. In Sec.~\ref{sec:exp:cot}, we evaluate the visual grounded CoT enabled by our implicit grounding approach. Finally, Sec.~\ref{sec:exp:abl} provides additional analysis and ablation studies of the model's 3D detection performance.

\begin{figure*}[t]
    \centering
    \includegraphics[width=1.0\linewidth]{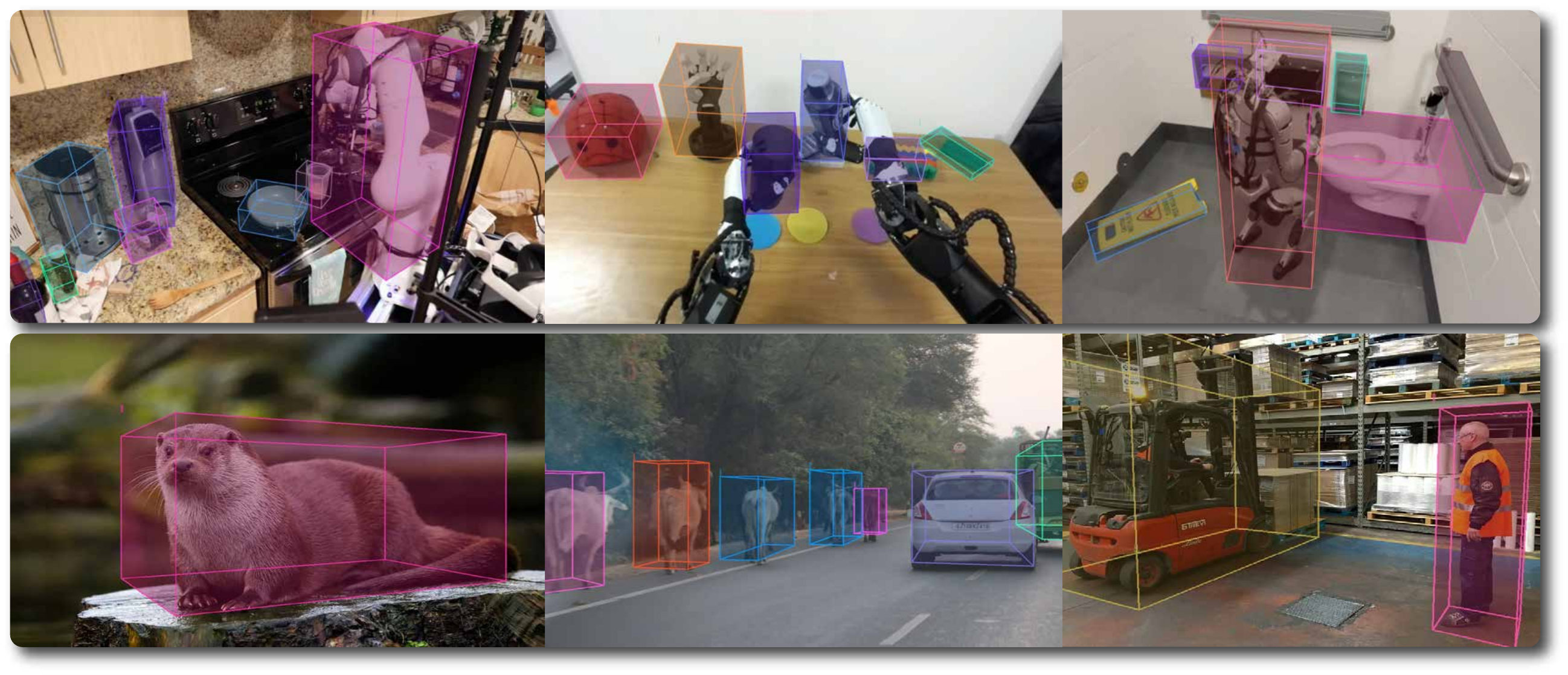}
    \vspace{-2.5em}
    \caption{Qualitative results on 3D object detection. Our model produces accurate 3D bounding boxes on in-the-wild samples.}
    \label{fig:qual}
    \vspace{-1.5em}
\end{figure*}


\subsection{Implementation Details}
\label{sec:exp:detail}
Our model is trained in two stages as detailed in following.

\noindent\textbf{Stage 1: Spatial Pretraining.}
The goal of this stage is to strengthen the model’s spatial understanding and 2D grounding capabilities, which later improves its 3D detection performance, as shown in our analysis. We initialize the visual encoder, projector, and LLM from NVILA-Lite 8B, while the spatial positional encoding module is newly initialized. Training is performed on a data mixture similar to SR-3D, augmented with 2D grounding data and region-to-3D detection data from Sec.~\ref{sec:data}. During this stage, we freeze the visual encoder and train the remaining modules.


\noindent\textbf{Stage 2: Detection CoT Finetuning.}
After pretraining, the model already possesses strong 2D grounding and basic 3D detection abilities. We then fine-tune it on CoT-oriented detection data, including detection data in CoT format (curated from Omni3D by first grounding in 2D and then predicting 3D boxes). Since the visual features are already well-formed after Stage 1, we fine-tune only the LLM to learn the reasoning and text-generation structure.


\subsection{3D Object Detection}
\label{sec:exp:det3d}
We evaluate our model on the Omni3D test set, following the benchmark protocol and hyperparameters used in DetAny3D. The Omni3D benchmark reports Average Precision (AP), where predictions are matched to ground-truth using 3D IoU with thresholds ranging from 0.05 to 0.50.

For comparison, we include both vision-specialist baselines (e.g., ImVoxelNet~\cite{rukhovich2022imvoxelnet}, Cube R-CNN~\cite{brazil2023omni3d}, OVMono3D~\cite{ovmono3d}, and DetAny3D~\cite{detany3d}) and VLM-based baselines (e.g., Qwen3VL-4B~\cite{qwen3vl} and Qwen3VL-8B~\cite{qwen3vl}. Our main results are shown in Table~\ref{tab:det3d} and Fig.~\ref{fig:qual}, where our model outperforms all VLM baselines. Compared with vision specialists, our model achieves competitive results overall and delivers notably better performance on indoor datasets.

We further analyze why existing VLMs perform worse on 3D detection. First, unlike our approach, they do not disentangle 3D detection into a two-step process—2D grounding followed by 3D box prediction. As we show in the analysis (Sec.~\ref{sec:exp:abl}), 2D grounding provides a stable geometric anchor that leads to more reliable and consistent 3D predictions.
Second, existing VLMs struggle with handling camera intrinsics. Qwen3VL is highly sensitive to input resolution, since pixel dimensions implicitly encode the focal length used in its geometric reasoning. This makes its 3D predictions unstable under changes in image size. VST~\cite{vst} partially addresses this by normalizing focal length in a manner similar to ours. However, it still requires FoVs to be passed as text prompts. Representing metric geometric parameters in textual form is difficult for the model to parse and integrate reliably, which limits its 3D understanding across scenes and camera setups.

Since our method explicitly separates 2D grounding from 3D prediction, we also evaluate 2D grounding performance on the Omni3D benchmark. As shown in Table~\ref{tab:indomain_ap2d}, our model exceeds region proposals generated by Cube R-CNN and the Qwen3-VL family. For Qwen3-VL models, which do not perform explicit 2D grounding, we evaluate using 2D boxes projected from their predicted 3D outputs.

\subsection{Visual Question Answering}
\label{sec:exp:vqa}
We investigate two key questions: (1) whether Stage 1 spatial pre-training effectively improves spatial reasoning performance, and (2) whether Stage 2 detection CoT finetuning negatively affects the model’s general VQA capabilities. We evaluate two variants of our model: one after spatial pre-training and one after CoT finetuning. The results are presented in Table~\ref{tab:spatial_benchmarks_condensed}. After spatial pre-training, the model shows a clear improvement on spatial-related VQA benchmarks, confirming the effectiveness of this stage. In contrast, Stage 2 finetuning focuses on learning the structure of CoT reasoning, and the results indicate that it does not significantly reduce general VQA performance. Most benchmarks remain similar to the Stage 1 model, suggesting that the model maintains strong general-purpose abilities.

\subsection{Implicit Grounding CoT}
\label{sec:exp:cot}
We aim to evaluate two aspects of our implicit grounding approach: (1) how accurate the grounding is, and (2) whether the grounding genuinely contributes to correct answers rather than producing hallucinated reasoning.

To study this, we evaluate our model on the MM-GCoT~\cite{wu2025grounded} benchmark, which provides three key metrics: answer accuracy (A-Acc), grounding accuracy (G-Acc), and answer–grounding consistency (Consist.). A-Acc measures the correctness of the textual answer. G-Acc follows the Acc@0.5 protocol, where a prediction is considered correct if its IoU with the ground-truth box exceeds 0.5. The consistency metric measures the percentage of predictions where both the answer and the grounding box are correct. We show results in Table~\ref{tab:mmgcot}, where our method outperforms baselines in all these metrics.

 To further evaluate the performance in spatial reasoning scenarios, we conduct experiments on BLINK-Depth using the same grounding-based CoT formulation. As shown in Table~\ref{fig:table_figure_shared}, our method surpasses prior Region-VLMs, which are typically strong on this benchmark but require manually annotated masks as input. In contrast, our model achieves higher performance while performing grounding automatically. We additionally provide qualitative examples demonstrating that our model can accurately localize tiny regions and successfully handle point-level areas.

\begin{table*}[!th]
\centering
\footnotesize
\scalebox{0.97}{
\setlength{\tabcolsep}{3.9pt}
\begin{tabular}{l ccc ccc ccc ccc}
\toprule
\multirow{2}{*}{{}} &
\multicolumn{3}{c}{{\small\textsc{Attribute}}} &
\multicolumn{3}{c}{{\small\textsc{Judgement}}} &
\multicolumn{3}{c}{{\small\textsc{Object}}} &
\multicolumn{3}{c}{{\small\textsc{Average}}} \\
\cmidrule(lr){2-4} \cmidrule(lr){5-7} \cmidrule(lr){8-10} \cmidrule(lr){11-13}
\multicolumn{1}{c}{} &
${\rm Acc_{A}} \uparrow$ & ${\rm Acc_{G}} \uparrow$ & ${\rm Cons.} \uparrow$ &
${\rm Acc_{A}} \uparrow$ & ${\rm Acc_{G}} \uparrow$ & ${\rm Cons.} \uparrow$ &
${\rm Acc_{A}} \uparrow$ & ${\rm Acc_{G}} \uparrow$ & ${\rm Cons.} \uparrow$ &
${\rm Acc_{A}} \uparrow$ & ${\rm Acc_{G}} \uparrow$ & ${\rm Cons.} \uparrow$ \\
\midrule
Qwen2.5-VL-7B~\cite{bai2025qwen2}(AF) & {73.6} & {72.5} & {59.8} & 87.9 & 56.3 & 51.5 & 57.8 & {64.1} & 59.1 & 73.1 & 64.3 & 56.8 \\
Qwen2.5-VL-7B~\cite{bai2025qwen2}(GF) & 48.8 & {82.6} & 45.7 & 80.6 & {72.8} & {62.4} & 26.7 & {62.3} & 32.6 & 52.0 & {72.6} & 46.9 \\
\midrule
LLaVA-7B~\cite{liu2024visual}(AF) & 68.6 & 9.2 & 8.8 & 83.0 & 11.2 & 11.5 & 58.4 & 9.1 & 9.9 & 70.0 & 9.8 & 10.1 \\
LLaVA-7B~\cite{liu2024visual}(GF) & 59.7 & 6.3 & 5.6 & 82.5 & 0.5 & 0.6 & 35.9 & 5.5 & 9.7 & 59.4 & 4.1 & 5.3 \\
LLaVA-GCoT-7B~\cite{wu2025grounded} & 72.8 & 66.7 & 56.1 & {88.3} & {61.7} & 56.9 & {62.3} & 61.7 & {61.3} & {74.5} & 63.3 & {58.1} \\
\rowcolor{myyellow!70}
\textbf{\model} (Ours) & \bf78.9 & \bf77.3 & \bf66.7 & \underline{85.0} & \bf79.6 & \bf70.4 & \bf71.1 & \bf65.7 & \bf66.1 & \bf78.3 & \bf74.2 & \bf67.7 \\
\bottomrule
\end{tabular}
}
\vspace{-1em}
\caption{Results on the MM-GCoT benchmark. “AF” and “GF” correspond to answer-first and grounding-first prompting settings. ${\rm Acc_{A}}$, ${\rm Acc_{G}}$, and ${\rm Cons.}$ refer to answer accuracy, grounding accuracy, and consistency between them.}
\label{tab:mmgcot}
\vspace{-2em}
\end{table*}

\subsection{Analysis and Ablation Study}
\label{sec:exp:abl}
\noindent\textbf{2D$\rightarrow$3D vs Direct 3D Prediction.}
As shown in Table~\ref{tab:abl-det}, first grounding the target region in 2D and then predicting its 3D bounding box leads to a clear improvement over direct 3D prediction. This two-step design is more vision-centric, as it explicitly forces the model to learn object-specific visual features before performing 3D reasoning. It also naturally decomposes the task into two subproblems—2D grounding and 3D inference—where the former benefits from significantly larger amounts of training data across generic detection and grounding datasets. Leveraging this abundant 2D supervision allows the model to establish stronger spatial priors, which in turn improves downstream 3D detection performance.

\begin{table}[!t]
\footnotesize
\centering

\scalebox{0.95}{
\setlength{\tabcolsep}{3pt}
\begin{tabular}{ccccccc}
\toprule
2D$\rightarrow$3D & PT & Cam & ${\rm AP^{sun}_{15}} \uparrow$ & ${\rm AP^{sun}_{3D}} \uparrow$ & ${\rm AP^{kit}_{15}} \uparrow$ & ${\rm AP^{kit}_{3D}} \uparrow$ \\
\midrule

-- & -- & -- & 30.19 & 20.27 & 10.08 & 6.22 \\

\checkmark & -- & -- & 42.29 & 29.87  & 15.61 & 10.03 \\

\checkmark & \checkmark & -- & 41.24 & 30.95 & 21.55 & 14.35 \\

\checkmark & \checkmark & \checkmark & \bf43.49 & \bf31.64 & \bf22.18 & \bf14.75 \\
\bottomrule
\end{tabular}
}
\vspace{-1em}
\caption{
Ablation study on the key components of \model. “PT” denotes pretraining, “2D$\rightarrow$3D” denotes 2D grounding followed by 3D prediction, and “Cam” denotes using normalized intrinsics.
}
\label{tab:abl-det}
\vspace{-2em}
\end{table}

\noindent\textbf{Do spatial pretraining help 3D detection?}
Table~\ref{tab:abl-det} further supports this assumption by showing that spatial pretraining noticeably improves performance in the outdoor domain. The Omni3D dataset is highly imbalanced~\cite{detany3d}, with far fewer outdoor training samples compared to indoor scenes. As a result, models trained from scratch struggle to generalize in outdoor settings. Spatial pretraining provides a strong remedy by injecting generic 2D spatial and grounding knowledge, enabling the model to better transfer its learned priors to the 3D detection task. This demonstrates that leveraging 2D supervision is especially beneficial when 3D data is limited or unevenly distributed.

\noindent\textbf{Effect of Intrinsic Normalization.}
Intrinsic normalization yields a modest, yet consistent improvement. Although its impact is smaller than the two factors discussed above, normalizing intrinsics helps reduce systematic biases when the model encounters cameras with different focal lengths. Without this normalization, the model may lead to small but noticeable localization offsets in the predicted 3D boxes.

\begin{figure}[!t]
    \centering
    \includegraphics[width=1.0\linewidth]{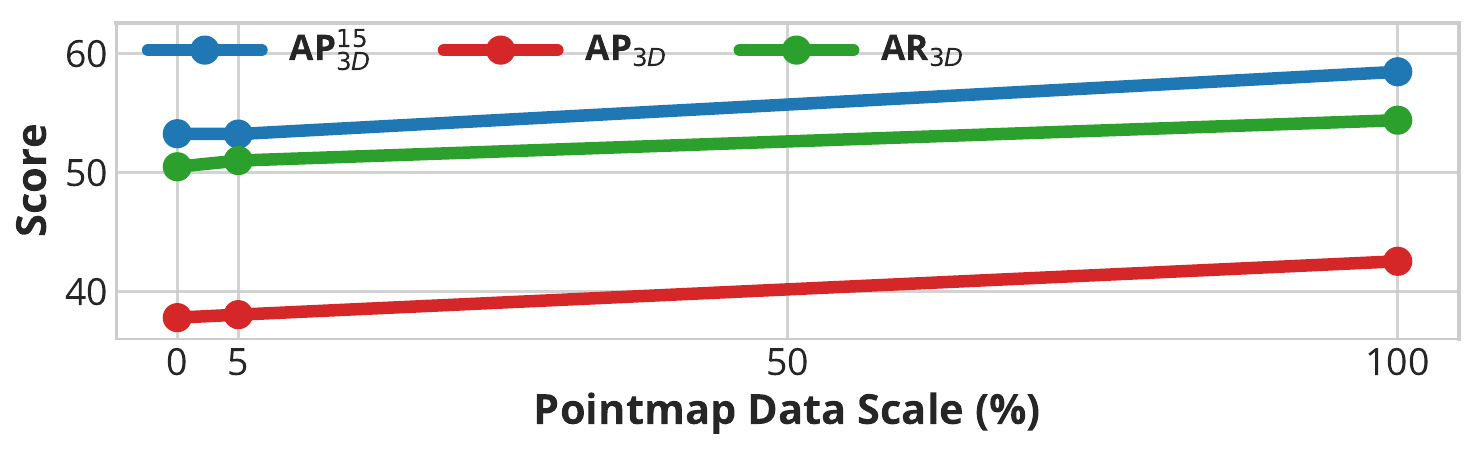}
    \vspace{-2em}
    \caption{Scaling behavior when increasing pointmap prediction data on SUN-RGBD. More pointmap supervision leads to better 3D detection performance.}
    \label{fig:scaling}
    \vspace{-2em}
\end{figure}

\noindent\textbf{Contribution of Pointmap Reconstruction.}
We further analyze the effect of pointmap reconstruction as an auxiliary task for 3D detection. This supervision strengthens the model’s ability to align region-level visual features with their corresponding 3D geometry. To isolate this effect from 2D grounding quality, we use ground-truth 2D boxes as our model input and evaluate only the 3D prediction. This separation is enabled by our disentangled pipeline and allows us to directly measure the reconstruction capability. As shown in Fig.~\ref{fig:scaling}, increasing the amount of pointmap supervision yields a clear scaling trend on SUN-RGBD: more pointmap data consistently improves 3D detection performance. 



\section{Related Work}
\label{sec:related}


\noindent\textbf{Spatial Vision Language Models.}
Recent work has rapidly expanded the spatial capabilities of VLMs across 2D grounding, monocular spatial reasoning, and multi-view 3D scene understanding~\cite{chen2024spatialvlm,ma2024spatialpin,cai2024spatialbot,yuan2024robopoint,ma20243dsrbench,tang2024sparkle,fu2025scene,song2024robospatial,xu2024llava,marsili2025visual,liu2025spatialcot,liao2024reasoning,yang2024thinking,man2024situational,linghu2024multi,embspatial,ray2024sat,wu2025spatialmllm,vst,spatialladder,yin2025spatial,hong20233d,ye2025omnivinci}. 
Representative 2D spatial VLMs such as SpatialVLM~\cite{chen2024spatialvlm}, SpatialPin~\cite{ma2024spatialpin}, VST~\cite{vst}, and SpatialLadder~\cite{spatialladder} focus on image-plane relations including relative position, direction, and distance using explicit spatial cues. 
SRGPT~\cite{srgpt} improves fine-grained single-view spatial perception by introducing a region branch for more precise region-level querying. 
SR-3D~\cite{cheng20253d} extends this idea by preserving and enabling multi-view spatial reasoning through a unified visual tokens space.
Other multi-view spatial VLMs~\cite{llava3d,video3dllm} incorporate 3D cues or cross-view alignment for scene reasoning. 
Despite this progress, implicit 2D grounding and monocular 3D grounding from a single image remain underexplored \cite{qwen3vl,vst,li2024seed}. 
In contrast, our approach jointly addresses both problems without requiring any spatial annotations at inference time.

\noindent\textbf{Monocular 3D Grounding.}
Traditional 3D object detection has long focused on single-dataset, closed-set scenarios~\cite{chen2016monocular,cheng20253d,li2024bevformer,liang2022bevfusion,liu2020smoke,rukhovich2022imvoxelnet,wang2021fcos3d,lin2022sparse4d,zhang2023monodetr}, achieving strong performance but suffering from poor generalization to new environments. Initial efforts to overcome this~\cite{brazil2023omni3d,li2024unimode} utilized multi-dataset training to create universal detectors. The Omni3D~\cite{brazil2023omni3d}, for instance, aggregated a wide variety of 3D datasets and proposed a universal model trained jointly on them. However, these models still confined to a predefined list of object classes seen during training. More recent work~\cite{wang2024ov,ovmono3d,detany3d} have turned their focus to the open-vocabulary setting. OVMono3D~\cite{ovmono3d} proposes a two-stage ``detect-then-lift" pipeline: it first employs an off-the-shelf 2D open-vocabulary detector~\cite{liu2024grounding} to generate 2D proposals, then feeds these region features into a specialized 3D head to regress 3D parameters. DetAny3D~\cite{detany3d} proposes a more integrated, promptable architecture that directly fuses features from 2D foundation models and uses 2D prompts, \eg, points, boxes, or text to query the model for 3D outputs. Instead of treating 3D grounding as a standalone detection task, GR3D predicts 3D boxes as part of a unified VLM framework that also includes dynamic implicit 2D grounding. This unified approach allows GR3D to leverage grounding as a key driver to enhance general spatial alignment and geometry-consistent reasoning, improving performance on both grounded and non-grounded spatial tasks.

\noindent\textbf{Thinking with Images.}
Our work is also realted to the recent line of “Thinking with Images’’ work~\cite{su2025thinking,yang2023mmreact, chen2025vitar,suris2023vipergpt, zhang2025thyme, fu2025refocus,hu2024sketchpad,vstar}. Different from these approaches, GR3D avoids explicit visual thought processes and external tools, offering a more efficient and unified design through implicit 2D grounding and native 3D reasoning within the VLM’s generative flow.

\section{Conclusion}
\label{sec:conclusion}

We introduced GR3D, a spatial VLM that integrates explicit 2D grounding, implicit grounding for CoT reasoning, and monocular 3D grounding within a single framework. By enabling the model to reference visual evidence during generation and by coupling region-grounded queries with 3D box prediction, GR3D decomposes spatial understanding into grounded 2D perception followed by 3D inference. GR3D delivers consistent gains across various benchmarks, showing that grounding serves as an effective inductive bias for better spatial understanding in VLMs.

\myparagraph{Acknowledgements.} This project was supported, in part, by NSF CAREER Award IIS-2240014, gifts from Amazon, Meta, and Qualcomm.

\clearpage
{
    \small
    \bibliographystyle{unsrt}
    \bibliography{main}
}

\clearpage
\setcounter{page}{1}
\setcounter{section}{0}
\maketitlesupplementary


\hypersetup{linkcolor=citecolor}

\begin{strip}
    \centering
    \includegraphics[width=\textwidth]{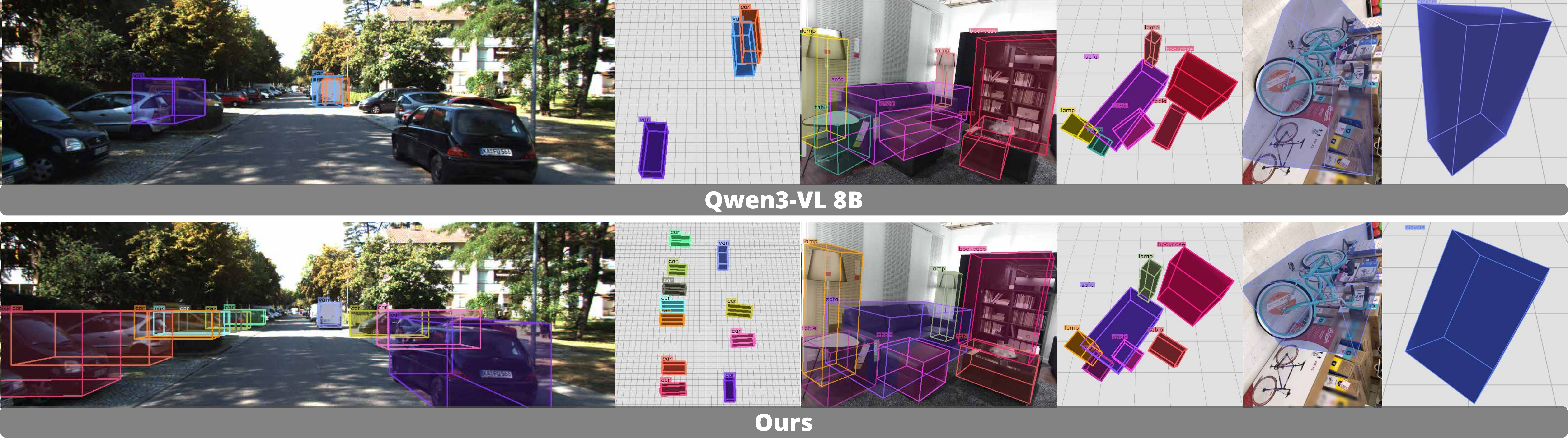}
    \captionof{figure}{
      Qualitative comparison on 3D object detection between our model and Qwen3-VL 8B~\cite{qwen3vl}. Our model produces more accurate 3D bounding boxes with fewer missed objects, demonstrating stronger spatial grounding and detection reliability.
    }
    \label{fig:sup:qual}
\end{strip}


\section*{Table of Contents}  
\startcontents[sections]
\printcontents[sections]{l}{1}{\setcounter{tocdepth}{3}}

\addtocontents{toc}{\protect\setcounter{tocdepth}{1}}

\section{More Results on 3D Detection}
\label{sec:sup:3d-det}
We show a qualitative comparison on 3D object detection between \method and Qwen3-VL-8B~\cite{qwen3vl}. As shown in Fig.~\ref{fig:sup:qual}, when multiple objects are present, \method produces clearly better results due to our detect-then-lift technique. For indoor scenes, \method also predicts 3D boxes with more accurate orientations compared to Qwen3-VL-8B.

\section{More Results on 2D Grounding}
\label{sec:sup:2d-ground}
Our approach decomposes 3D detection into two steps: first grounding the target in 2D, then predicting its 3D properties based on the grounded region. Because accurate 2D grounding is essential for the first step, we evaluate our model on two grounding benchmarks. We first report results on RefSpatial~\cite{zhou2025roborefer}, a benchmark designed for spatial referring that includes queries about vacant regions, spatial relations (e.g., ``left of'', ``between''), and fine-grained spatial logic. As shown in Table~\ref{tab:sup:refspatial}, our model achieves strong spatial referring performance and outperforms several baselines, including RoboRefer~\cite{zhou2025roborefer}, demonstrating its ability to reason about complex spatial relations in 2D. We further evaluate on the widely used RefCOCO, RefCOCO+, and RefCOCOg datasets~\cite{kazemzadeh2014referitgame, mao2016generation} to measure general referring capability. These benchmarks contain diverse referring expressions involving object names, attributes, and relational descriptions. Results in Table~\ref{tab:sup:refcoco} show that \method performs comparably to vision-specialized models and is on par with top VLMs such as InternVL-3.5~\cite{internvl35} or Qwen2.5-VL~\cite{bai2025qwen2}, confirming that our strong 2D grounding ability generalizes well to both spatial and standard referring benchmarks.

\begin{table}[!ht]
\footnotesize
\centering
\scalebox{0.97}{
\setlength{\tabcolsep}{7pt}
\begin{tabular}{l ccc}
\toprule
Method
& \textsc{Location} & \textsc{Placement} & \textsc{Unseen} \\
\midrule

Gemini-2.5-Pro~\cite{gemini_2_5}
& 46.9 & 24.2 & 27.1 \\

SpaceLLaVA-13B~\cite{chen2024spatialvlm}
& 5.8 & 4.3 & 4.0 \\

RoboPoint-13B~\cite{yuan2024robopoint}
& 22.8 & 9.2 & 8.4 \\

Molmo-7B~\cite{deitke2025molmo}
& 21.9 & 12.8 & 12.2 \\

Molmo-72B~\cite{deitke2025molmo}
& 45.7 & 14.7 & 21.2 \\

RoboBrain-2.0-7B~\cite{RoboBrain2.0}
& 36.0 & 29.0 & 32.5 \\

RoboRefer-8B~\cite{zhou2025roborefer}
& 52.0 & 53.0 & 37.7 \\

\rowcolor{myyellow!70}
\textbf{\model}
& \bf63.0 & \bf50.0 & \bf41.5 \\
\bottomrule
\end{tabular}
}
\caption{
Performance comparison on RefSpatial~\cite{zhou2025roborefer}.
}
\label{tab:sup:refspatial}
\end{table}

\begin{table*}[!t]
    \footnotesize
    \centering
    \setlength\tabcolsep{6.5pt} 
    \scalebox{0.97}{
    \begin{tabular}{lccccccccccc}
    & 
    \rotatebox{55}{Obj. Count} &
    \rotatebox{55}{Abs. Dist.} &
    \rotatebox{55}{Obj. Size} & 
    \rotatebox{55}{Room Size} &
    \rotatebox{55}{Rel. Dist.} &
    \rotatebox{55}{Rel. Dir.} &
    \rotatebox{55}{Route Plan} &
    \rotatebox{55}{Appr. Order} \\
    Methods & \multicolumn{4}{c}{\cellcolor{orange!10}Quantitative} & \multicolumn{4}{c}{\cellcolor{yellow!10}Qualitative} & {\cellcolor{red!10}Avg.} \\
    \midrule
    \color[HTML]{969696}Random & \color[HTML]{969696}- & \color[HTML]{969696}- & \color[HTML]{969696}- & \color[HTML]{969696}- & \color[HTML]{969696}25.0 & \color[HTML]{969696}36.1 & \color[HTML]{969696}28.3 & \color[HTML]{969696}25.0 & \color[HTML]{969696}- \\
    \color[HTML]{969696}Human Level\textsuperscript{\dag} & \color[HTML]{969696}94.3 & \color[HTML]{969696}47.0 & \color[HTML]{969696}60.4 & \color[HTML]{969696}45.9 & \color[HTML]{969696}94.7 & \color[HTML]{969696}95.8 & \color[HTML]{969696}95.8 & \color[HTML]{969696}100 & \color[HTML]{969696}79.2 \\
    \midrule
    \rowcolor{navyblue!5}
    \multicolumn{1}{l}{\textcolor{black}{\bf\emph{Proprietary Models (API)}}} & & & & & & & & & \\
    GPT-4o~\citep{openai2024gpt4o} & 46.2 & 5.3 & 43.8 & 38.2 & 37.0 & 41.3 & 31.5 & 28.5 & 34.0 \\
    Gemini-1.5 Flash~\citep{team2024gemini} & 49.8 & 30.8 & 53.5 & 54.4 & 37.7 & 41.0 & 31.5 & 37.8 & 42.1 \\
    Gemini-1.5 Pro~\citep{team2024gemini} & 56.2 & 30.9 & 64.1 & 43.6 & 51.3 & 46.3 & 36.0 & 34.6 & 45.3 \\
    \midrule
    \rowcolor{navyblue!5}
    \multicolumn{1}{l}{\textcolor{black}{\bf\emph{Open-source Models}}} & & & & & & & & & \\
    InternVL2-2B~\cite{chen2024internvl} & 24.9 & 22.0 & 35.0 & 33.8 & {44.2} \\
    InternVL2-8B~\citep{chen2024internvl} & 31.3 & {29.0} & 48.9 & {44.2} & 38.0 & 33.4 & 28.9 & 46.4 & 37.5  \\
    InternVL2-40B~\citep{chen2024internvl} & 41.3 & 26.2 & 48.2 & 27.5 & 47.6 & 32.7 & 27.8 & 44.7 & 37.0  \\
    LongVILA-8B~\citep{xue2024longvila} & 29.1 & 9.1 & 16.7 & 0.0 & 29.6 & 30.7 & 32.5 & 25.5 & 21.6  \\
    VILA-1.5-8B~\cite{lin2024vila} & 17.4 & 21.8 & 50.3 & 18.8 & 32.1 & 34.8 & 31.0 & 24.8 & 28.9  \\
    VILA-1.5-40B~\cite{lin2024vila} & 22.4 & 24.8 & 48.7 & 22.7 & 40.5 & 25.7 & 31.5 & 32.9 & 31.2  \\
    LongVA-7B~\citep{zhang2024long} & 38.0 & 16.6 & 38.9 & 22.2 & 33.1 & 43.3 & 25.4 & 15.7 & 29.2  \\
    LLaVA-Video-7B~\citep{llavanext} & 48.5 & 14.0 & 47.8 & 24.2 & {43.5} & 42.4 & 34.0 & 30.6 & 35.6  \\
    LLaVA-Video-72B~\citep{llavanext} & {48.9} & 22.8 & 57.4 & 35.3 & 42.4 & 36.7 & {35.0} & {48.6} & 40.9  \\
    LLaVA-OneVision-7B~\citep{li2025llavaonevision} & 47.7 & 20.2 & 47.4 & 12.3 & 42.5 & 35.2 & 29.4 & 24.4 & 32.4  \\
    LLaVA-OneVision-72B~\citep{li2025llavaonevision} & 43.5 & 23.9 & {57.6} & 37.5 & 42.5 & 39.9 & 32.5 & 44.6 & 40.2  \\
    SR-3D-8B & {54.9} & 53.8 & 74.5 & 65.1 & 63.5 & 81.8 & 33.5 & 75.9 & 62.9 \\
    \rowcolor{myyellow!70}
    \textbf{\method-8B} & \bf69.6 & \bf55.2 & \bf76.8 & \bf65.6 & \bf70.5 & \bf86.3 & \bf35.5 & \bf81.2 & \bf67.6 \\
    \bottomrule
    \end{tabular}
} 
    \caption{We finetune our model on multi-view datasets~\cite{azuma2022scanqa,wang2024embodiedscan} following SR-3D~\cite{cheng20253d}, and then evaluate multi-view global spatial scene understanding on VSI-Bench~\cite{yang2024thinking}. Methods marked with \textsuperscript{\dag} are evaluated on the \texttt{Tiny} subset. \method outperforms all state-of-the-art baselines, demonstrating strong spatial recognition capability.}
    \label{tab:sup:3d_vsibench}
\end{table*}

\begin{table}[!ht]
\footnotesize

    \centering
    \scalebox{0.97}{
    \setlength{\tabcolsep}{1.2pt} 
    \begin{tabular}{l ccc ccc cc}
    \toprule
    \multirow{2}{*}{Model Name}
    & \multicolumn{3}{c}{\textsc{RefCOCO}}
    & \multicolumn{3}{c}{\textsc{RefCOCO+}}
    & \multicolumn{2}{c}{\textsc{RefCOCOg}} \\
    \cmidrule(lr){2-4} \cmidrule(lr){5-7} \cmidrule(lr){8-9}
    & {val} & {testA} & {testB}
    & {val} & {testA} & {testB}
    & {val} & {test} \\
    \midrule
    \rowcolor{navyblue!5}
    \multicolumn{1}{l}{\textcolor{black}{\bf\emph{Vision Specialists}}} & & & & & & & & \\
    Grounding-DINO-L~\citep{liu2024grounding} & 90.6 & 93.2 & 88.2 & 82.8 & 89.0 & 75.9 & 86.1 & 87.0 \\
    UNINEXT-H~\citep{yan2023universal}               & 92.6 & 94.3 & 91.5 & 85.2 & 89.6 & 79.8 & 88.7 & 89.4 \\
    ONE-PEACE~\citep{wang2023one}             & 92.6 & 94.2 & 89.3 & 88.8 & 92.2 & 83.2 & 89.2 & 89.3 \\
    \midrule
    \rowcolor{navyblue!5}
    \multicolumn{1}{l}{\textcolor{black}{\bf\emph{Vision Language Models}}} & & & & & & & & \\
    InternVL3-1B~\cite{internvl3}    & 85.8 & 90.1 & 81.7 & 76.6 & 84.1 & 69.2 & 82.8 & 82.6 \\
    InternVL3.5-1B~\cite{internvl35}      & 85.4 & 89.7 & 80.2 & 77.7 & 85.5 & 69.5 & 81.9 & 81.6 \\
    InternVL3-2B~\cite{internvl3}    & 89.8 & 92.6 & 86.4 & 84.0 & 89.2 & 76.5 & 87.6 & 87.2 \\
    InternVL3.5-2B~\cite{internvl35}   & 88.7 & 91.6 & 84.8 & 82.7 & 88.4 & 76.6 & 85.6 & 85.5 \\
    Qwen2.5-VL-3B~\cite{bai2025qwen2}     & 89.1 & 91.7 & 84.0 & 82.4 & 88.0 & 74.1 & 85.2 & 85.7 \\
    Shikra-7B~\citep{chen2023shikra}        & 87.0 & 90.6 & 80.2 & 81.6 & 87.4 & 72.1 & 82.3 & 82.2 \\
    CogVLM-G~\citep{wang2023cogvlm} & 92.8 & 94.8 & 89.0 & 88.7 & 92.9 & 83.4 & 89.8 & 90.8 \\
    Qwen2-VL-7B~\cite{wang2024qwen2}      & 91.7 & 93.6 & 87.3 & 85.8 & 90.5 & 79.5 & 87.3 & 87.8 \\
    Qwen2.5-VL-7B~\cite{bai2025qwen2}     & 90.0 & 92.5 & 85.4 & 84.2 & 89.1 & 76.9 & 87.2 & 87.2 \\
    TextHawk2~\citep{yu2024texthawk}       & 91.9 & 93.0 & 87.6 & 86.2 & 90.0 & 80.4 & 88.2 & 88.1 \\
    InternVL3.5-8B~\cite{internvl35}   & 92.4 & 94.7 & 88.7 & 87.9 & 92.4 & 82.4 & 89.6 & 89.4 \\
    \rowcolor{myyellow!70}
    \textbf{\model} & 91.8 & 94.5 & 88.8 & 87.5 & 91.4 & 81.0 & 89.5 & 89.7 \\
    \bottomrule
    \end{tabular}
    }
    \caption{
    We evaluate \method’s 2D grounding on RefCOCO, RefCOCO+, and RefCOCOg~\cite{kazemzadeh2014referitgame, mao2016generation}. Baseline numbers are taken from \cite{wang2024qwen2, internvl35}. \method achieves grounding accuracy comparable to vision specialists~\cite{liu2024grounding,yan2023universal,wang2023one} models and performs on par with top VLMs such as InternVL3.5~\cite{internvl35}.
    }
    \label{tab:sup:refcoco}
\end{table}

\section{More Results on Multi-View Understanding}
\label{sec:sup:multiview}

\subsection{Multi-View Extension of Our Framework}

Our framework naturally extends from single-view to multi-view settings through a unified spatial embedding design similar to SR-3D~\cite{cheng20253d}. All image tokens, regardless of the view they originate from, are mapped into the same spatial feature space using depth-based and pixel-based positional cues. This allows the model to maintain consistent geometric relationships across views without requiring explicit point cloud reconstruction or global world coordinates.

For multi-view inputs, the first view is processed exactly as in the single-view case and is treated as the reference frame. Unlike SR-3D, which assumes a global world coordinate system and expresses all views in that space, our approach keeps everything in the coordinate frame of the first camera. Each additional view is transformed into this reference coordinate system using its intrinsics and extrinsics, so all depth-derived 3D locations and pixel-coordinate cues are expressed in the same spatial frame. After this transformation, tokens from different views that observe the same physical point occupy nearby positions in the unified embedding space. This allows the model to reason about 3D structure, occlusion, and cross-view consistency directly from the spatial tokens.

\subsection{Results on VSI-Bench}

To validate this design, we finetune our stage-1 model on multi-view datasets~\cite{azuma2022scanqa,wang2024embodiedscan} following SR-3D~\cite{cheng20253d}, and then evaluate multi-view global spatial scene understanding on VSI-Bench~\cite{yang2024thinking}. As shown in Table~\ref{tab:sup:3d_vsibench}, \method achieves strong performance with an average score of 67.6 and surpasses all state-of-the-art baselines, showing that our method can effectively handle multi-view inputs.

\subsection{Results on ScanRefer, ScanQA, MMSI, SPAR}
\label{sec:sup:3d-ground}

To further evaluate the 3D grounding capabilities of \method on multi-view datasets, we conduct studies leveraging ScanRefer~\cite{chen2020scanrefer} benchmark. However, ScanRefer assumes access to a pre-aligned world coordinate space, which is not directly compatible with the settings of Qwen3-VL~\cite{qwen3vl} and ours. We therefore adapt ScanRefer into a frame/2D box grounding followed by 3D detection in the camera coordinate space, and compare against Qwen3-VL under the same input conditions. Our method outperforms Qwen3-VL-8B and is competitive with methods that use pre-aligned 3D input. We also report results on ScanQA~\cite{azuma2022scanqa}, MMSI-Bench~\cite{yang2025mmsi} and SPAR-Bench~\cite{spar}, showing consistent improvements.

\begin{table}[!h]
    \scriptsize
    \centering
    \begin{minipage}{0.23\textwidth}
        \centering
        \scalebox{0.95}{
        \setlength{\tabcolsep}{1pt}
        \begin{tabular}{lccccc}
        \toprule
        & \multicolumn{2}{c}{\textsc{Scanrefer}} & \multicolumn{3}{c}{\textsc{Scanqa}} \\
        \cmidrule(lr){2-3} \cmidrule(lr){4-6}
        & @0.25 & @0.5 & B4 & C & EM \\
        \midrule
        SPAR & 48.8 & 43.1 & 15.3 & 90.7 & 27.7 \\
        3D-LLaVA & 51.2 & 40.6 & 17.1 & 92.6 & - \\
        Video-3D LLM  & 58.1 & 51.7 & 16.2 & 102.1 & 30.1 \\
        \midrule
        Qwen3-VL-8B & 37.7 & 33.2 & - & - & - \\
        \rowcolor{myyellow!70}
        \textbf{GR3D-8B}  & {52.0} & {46.1} & {18.1} & {105.1} & {29.2}\\
        \bottomrule
        \end{tabular}
        }
    \end{minipage}
    \hfill 
    \begin{minipage}{0.23\textwidth}
        \centering
        \scalebox{0.95}{
        \setlength{\tabcolsep}{2.5pt}
        \begin{tabular}{lcc}
        \toprule
        & \textsc{MMSI} & \textsc{SPAR} \\
        \midrule
        GPT-4o & 30.3 & 38.1 \\
        InternVL2.5-8B  & 28.7  & 36.3 \\
        Qwen2.5-VL-7B   & 25.9  & 33.1 \\
        Qwen3-VL-8B  & 31.1  &  39.8 \\
        \midrule
        NVILA-8B  & 28.1  & 34.1 \\
        SR-3D-8B  & 25.8  & 32.1 \\
        \rowcolor{myyellow!70}
        \textbf{GR3D-8B} & 29.2 & 43.7 \\
        \bottomrule
        \end{tabular}
        }
    \end{minipage}
    \caption{Performance comparison on ScanRefer~\cite{chen2020scanrefer}, ScanQA~\cite{azuma2022scanqa}, MMSI-Bench~\cite{yang2025mmsi}, and SPAR-Bench~\cite{spar}.}
    \label{tab:multi_view_3d_ground}
\end{table}

\section{Implementation Details}
\label{sec:sup:imp-details}

\subsection{Model Architecture}
Following NVILA-Lite, we use SigLIP as the vision encoder with an input resolution of 448 and a patch size of 14, paired with a Qwen-2-7B~\citep{wang2024qwen2} LLM backbone. For training the stage-1 model, we follow SR-3D and enable dynamic tiling with up to 12 tiles per image. We also adopt SR-3D's dynamic tiling region extractor, which provides a larger effective receptive field for regions and improves the model's ability to handle small objects. During the first stage, the vision encoder is frozen and only the remaining modules are trained. For the second CoT detection stage, the LLM is fine-tuned to learn the reasoning structure and the autoregressive 3D prediction format.

\subsection{Training Hyper-parameters}
Both stages use the same optimization schedule: a warmup ratio of 0.03 and a cosine learning-rate scheduler. In the stage-1 stage, we train all non-visual modules with AdamW and a base learning rate of $5\times10^{-5}$, while keeping the SigLIP encoder frozen. The second CoT detection stage fine-tunes only the Qwen-2-7B LLM with a smaller learning rate of $1.5\times10^{-5}$ to stabilize chain-of-thoughts text generation. Training the stage-1 model takes approximately 4 days on 8 nodes of A100 servers, while the second stage takes about 4 hours on the same compute setup.

\begin{table}[!h]
    \scriptsize\centering
    \begin{tabular}{rp{150pt}}
        \toprule
        \multicolumn{2}{l}{\textit{\textbf{Stage-1 Data}}} \\
        \midrule
        Hybrid     & ShareGPT4V-SFT, Molmo, The Cauldron, Cambrian, LLaVA-OneVision \\
        \midrule
        Captioning & MSR-VTT, Image Paragraph Captioning, ShareGPT4V-100K \\
        \midrule
        Reasoning  & CLEVR, NLVR, VisualMRC \\
        \midrule
        Document & DocVQA, UniChart-SFT, ChartQA \\ 
        \midrule
        OCR & TextCaps, OCRVQA, ST-VQA, POIE, SORIE, SynthDoG-en, TextOCR-GPT4V, ArxivQA, LLaVAR \\
        \midrule
        General VQA & ScienceQA, VQAv2, ViQuAE, Visual Dialog, GQA, Geo170K, LRV-Instruction, RefCOCO, GeoQA, OK-VQA, TabMVP, EstVQA \\
        \midrule
        Diagram \& Dialogue & DVQA, AI2D, Shikra, UniMM-Chat \\
        \midrule
        Instruction & LRV-Instruction, SVIT, MMC-Instruction, MM-Instruction \\
        \midrule
        Text-only & FLAN-1M, MathInstruct, Dolly, GSM8K-ScRel-SFT \\
        \midrule
        Knowledge & WordART, WIT, STEM-QA \\
        \midrule
        Medical & PathVQA, Slake, MedVQA \\
        \midrule
        Region & RegionGPT \\
        \midrule
        Spatial \& 2D Grounding & RefCOCO, MGrounding, Molmo, Groma, SpatialRGPT, RefSpatial, SAT, EmbSpatial, DepthLM \\
        \midrule
        Detection & Omni3D, EmbodiedScan \\
        \midrule
       \multicolumn{2}{l}{\textit{\textbf{Stage-2 Data}}} \\
        \midrule
        Detection & Omni3D-CoT \\
        \midrule
        Spatial & RefSpatial-CoT, MMG-CoT, EmbSpatial-CoT, VisCoT \\
        \bottomrule
    \end{tabular}
    \caption{Data recipe for training \method.}
    \label{tab:recipe:data}
\end{table}

\subsection{Data Composition}

\begin{figure*}[!t]
    \centering
    \includegraphics[width=\linewidth]{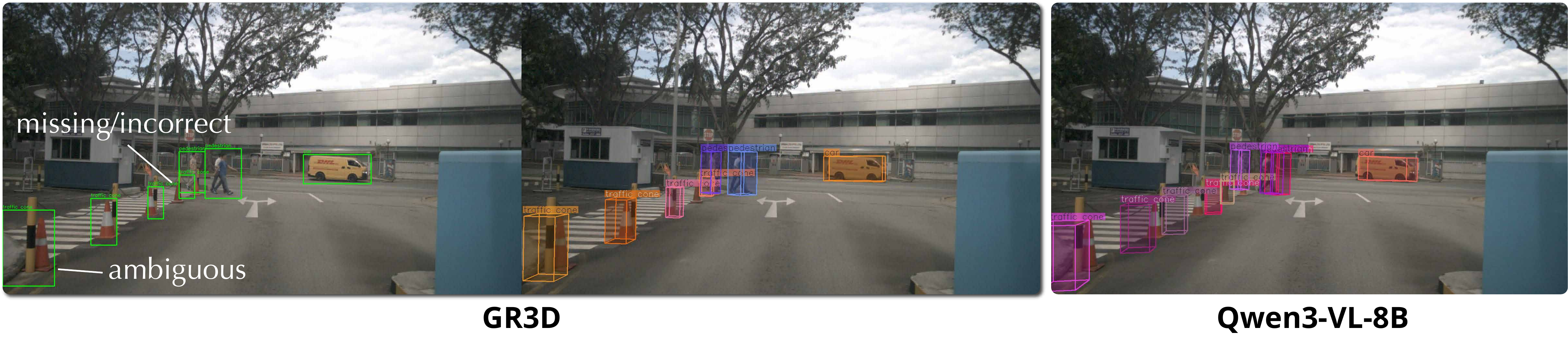}
    \caption{A failure case analysis of \method compared to Qwen3-VL-8B.}
    \label{fig:failure_case}
\end{figure*}

The data composition for both training stages is summarized in Table~\ref{tab:recipe:data}. Most of our training data follow NVILA's data recipe, though we use only a subset due to computational constraints. Part of the spatial data is inherited from SR-3D, while many of the 2D grounding datasets are newly introduced to the model and trained for the first time on our weights. For the 3D detection data used in stage~1, we follow DetAny3D's filtering rules on Omni3D to select high-quality training objects, and convert each scene into multi-turn conversations with up to 10 rounds. For the CoT detection data used in stage~2, we construct multi-object reasoning sequences by selecting up to 20 objects for each target.

\section{More Related Work}
\label{sec:sup:think-w-images}
A related line of research, recently formalized as \textit{Thinking with Images}~\cite{su2025thinking}, focuses on improving complex VLM reasoning by decomposing problems into explicit, intermediate steps, treating vision as a dynamic workspace. Many such methods act as ``commanders" orchestrating external visual tools~\cite{yang2023mmreact, chen2025vitar} or as ``visual programmers" that generate code for custom analysis and edits~\cite{suris2023vipergpt, zhang2025thyme, fu2025refocus}. Others generate intermediate visual representations to guide reasoning, often called Visual Chain of Thought (V-CoT)~\cite{zhang2023visual}. These V-CoT methods may interleave text with explicit visual groundings~\cite{fan2025grit}, sketch visual artifacts~\cite{hu2024sketchpad}, generate subgoal images for robotics~\cite{zhao2025cotvla}, or perform planning entirely through visual state sequences~\cite{xu2025visualplanning}. 
While these methods enhance transparency and performance on complex tasks, they still focus on 2D image space, rely on coarse region-selection cues or external tools, and rarely integrate these reasoning steps with a 3D spatial framework. 
In contrast, our GR3D framework bypasses the need for an explicit, step-by-step visual thought process. It achieves a more seamless integration by performing implicit 2D grounding and unified 3D reasoning natively within the VLM's generative flow.

\section{Discussions}
\label{sec:sup:discussions}

\subsection{Standard VLM without PE}
Our method can be applied to standard VLMs, but 3D priors further improve performance. Using positional embeddings, Omni3D mAP (averaged over 6 datasets) improves from 22.9 to 25.4 compared to a standard VLM without positional embeddings, showing their benefit as a simple and effective 3D prior.

\subsection{Hallucinations}

We do not observe frequent hallucinated 2D boxes. The main failures are missing or ambiguous 2D grounding, which leads to incorrect predictions. We show an example above and compare them with Qwen3-VL-8B~\cite{qwen3vl}.

\subsection{Effect of Intrinsic Estimation Errors}
The effect of intrinsic normalization is modest. Since the normalization only determines the resolution size, it does not require highly accurate intrinsic estimates. In practice, off-the-shelf intrinsic estimators are sufficiently accurate: using GeoCalib for focal length prediction on Omni3D results in only a 1.2 mAP drop (averaged over 6 datasets).

\subsection{CoT Data Robustness}
We conduct an ablation study on the impact of data quality by training with a noisier corpus, which leads to performance drops from 74.2 to 62.8 on MM-GCoT's grounding accuracy. Human evaluation on 200 randomly sampled instances from the filtered corpus shows that 95.5\% of the generated bounding boxes are accurate.

\subsection{Latency Analysis}
We implement multimodal prefix caching to ensure that the inference pipeline runs at a speed comparable to standard autoregressive generation. For Region Insertion, the process only extracts relevant areas from already encoded image features and passes them through a lightweight MLP projector, without re-encoding the image. We provide a latency analysis that compares our method with other baselines, tested on the same input image using a single A100 GPU. Our model is fastest among VLMs due to a more efficient dynamic tiling–based vision encoder (vs. AnyRes). The additional cost per inserted region is only 0.01 s, which is a small fraction of the total 2.7 s inference time.

\begin{table}[!h]
\scriptsize
    \centering
    \scalebox{0.97}{
    \setlength{\tabcolsep}{8.5pt}
    \begin{tabular}{lccccc}
    \toprule
    & DetAny3D & VST-7B & Qwen3-VL-8B & \textbf{\model} \\
    \midrule
    Latency (s) & 0.98 & 2.76 & 3.23 & 2.72 \\
    \bottomrule
    \end{tabular}
    }
\end{table}

\subsection{Limitations}
\label{sec:sup:limitations}
Our approach has two main limitations. First, the inference speed is slower compared to vision specialists. This is mainly due to the use of a large language model backbone, our two-stage ``2D grounding first'' pipeline, and the fact that 3D bounding boxes are generated autoregressively as text tokens, all of which introduce additional latency. Second, current 3D detection datasets are still limited. Popular datasets such as Omni3D cover only a narrow set of environments, camera configurations, and object categories, which restricts the diversity and scale of 3D supervision our model can learn from. As a result, further progress will benefit from larger and more diverse 3D datasets with broader scene coverage and richer object annotations.

\end{document}